\documentclass[runningheads]{llncs}
\usepackage{graphicx}
\usepackage{amsfonts}
\usepackage{amsmath}

\begin{document}
\title{Deep Reinforcement Learning for L3 Slice Localization in Sarcopenia Assessment}
\titlerunning{Deep Reinforcement Learning for Slice Localization}
%
%
\author{Othmane Laousy\inst{1,2,3} \and
Guillaume Chassagnon\inst{2} \and
Edouard Oyallon\inst{4}
\and
Nikos Paragios\inst{5}
\and
Marie-Pierre Revel\inst{2}
\and
Maria Vakalopoulou\inst{1,3}}
\authorrunning{O. Laousy et al.}
%
\institute{CentraleSupélec, Paris-Saclay University, Mathématiques et Informatique pour la
Complexité et les Systèmes, Gif-sur-Yvette, France.
\and
Radiology Department, Hôpital Cochin AP-HP, Paris, France
\and
Inria Saclay, Gif-sur-Yvette, France
\and
Centre National de Recherche Scientifique, LIP6, Paris, France
\and
Therapanacea, Paris, France
}

\maketitle              

\begin{abstract}
Sarcopenia is a medical condition characterized by a reduction in muscle mass and function. A quantitative diagnosis technique consists of localizing the CT slice passing through the middle of the third lumbar area (L3) and segmenting muscles at this level. In this paper, we propose a deep reinforcement learning method for accurate localization of the L3 CT slice. Our method trains a reinforcement learning agent by incentivizing it to discover the right position. Specifically, a Deep Q-Network is trained to find the best policy to follow for this problem. Visualizing the training process shows that the agent mimics the scrolling of an experienced radiologist.  Extensive experiments against other state-of-the-art deep learning based methods for L3 localization prove the superiority of our technique which performs well even with a limited amount of data and annotations.

\keywords{L3 slice  \and CT slice localization \and Deep Reinforcement Learning \and Sarcopenia}
\end{abstract}

\section{Introduction}
Sarcopenia corresponds to muscle atrophy which may be due to ageing, inactivity, or disease. The decrease of skeletal muscle is a good indicator of the overall health state of a patient~\cite{sarcopeniahealth}. In oncology, it has been shown that sarcopenia is linked to outcome in patients treated by chemotherapy~\cite{lee2018skeletal,sarcochemo}, immunotherapy~\cite{sarcoimmuno}, or surgery~\cite{sarcosurgery}. 
There are multiple definitions of sarcopenia~\cite{cruz2019sarcopenia,santilli2014clinical} and consequently multiple ways of assessing it. On CT imaging, the method used is based on muscle mass quantification. Muscle mass is most commonly assessed at a level passing through the middle of the third lumbar vertebra area (L3), which has been found to be representative of the body composition~\cite{zopfs}. After manual selection of the correct CT slice at the L3 level, segmentation of muscles is performed to calculate the skeletal muscle area~\cite{measurement}.
In practice, the evaluation is tedious, time-consuming, and rarely done by radiologists, highlighting the need for an automatic diagnosis tool that could be integrated into clinical practice. Such automated measurement of muscle mass could be of great help for introducing sarcopenia assessment in daily clinical practice.

Muscle segmentation and quantification on a single slice have been thoroughly addressed in multiple works using simple 2D U-Net like architectures~\cite{blanc2020abdominal,castiglione2021automated}.
Few works, however, focus on L3 slice detection. The main challenges for solving this task rely on the inherent diversity in patient's anatomy, the strong resemblance between vertebrae, the variability of CT fields of view as well as their acquisition and reconstruction protocols.

The most straightforward approach to address L3 localization is by investigating methods for multiple vertebrae labeling in 3D images using detection~\cite{suzani} or even segmentation algorithms~\cite{Payer2020}. Such methods require a substantial volume of annotations and are computationally inefficient when dealing with the entire 3D CT scan. In fact, even if our input is 3D, a one-dimensional output as the $z$-coordinate of the slice is sufficient to solve the L3 localization problem. 

In terms of L3 slice detection, the closest methods leverage deep learning~\cite{belharbi2017spotting,kanavati2018automatic} and focus on training simple convolutional neural networks (CNN). These techniques use maximal intensity projection (MIP), where the objective is to project voxels with maximal intensity values into a 2D plane. Frontal view MIP projections contain enough information towards the body and vertebra's bone structure differentiation. On the sagittal view, restricted MIP projections are used to focus solely on the spinal area.
 In~\cite{belharbi2017spotting} the authors tackle this problem through regression, training the CNN with parts of the MIP that contain the L3 vertebra only. More recently, in~\cite{kanavati2018automatic} a UNet-like architecture (L3UNet-2D) is proposed to draw a 2D confidence map over the position of the L3 slice.

In this paper, we propose a reinforcement learning algorithm for accurate detection of the L3 slice in CT scans, automatizing the process of sarcopenia assessment. The main contribution of our paper is a novel formulation for the problem of L3 localization, exploiting different deep reinforcement learning (DRL) schemes that boost the state of the art for this challenging task, even on scarce data settings. Moreover, in this paper we demonstrate that the use of 2D approaches for vertebrae detection provides state of the art results compared to 3D landmark detection methods, simplifying the problem, reducing the search space and the amount of annotations needed.
To the best of our knowledge, this is the first time a reinforcement learning algorithm is explored on vertebrae slice localization, reporting performances similar to medical experts and opening new directions for this challenging task.

\section{Background}
Reinforcement Learning is a fundamental tool of machine learning which allows dealing efficiently with the exploration/exploitation trade-off~\cite{sutton2018reinforcement}. Given state-reward pairs, a reinforcement learning agent can pick actions to reach unexplored states or increase its accumulated future reward. Those principles are appealing for medical applications because they imitate a practitioner's behavior and self-learn from experience based on ground-truth. One of the main issues of this class of algorithm is its sample complexity:  a large amount of interaction with its environment is needed before obtaining an agent close to an optimal state~\cite{tarbouriech2020no}. However, those techniques were recently combined with deep learning approaches, which efficiently addressed this issue~\cite{mnih2013playing} by incorporating priors based on neural networks. In the context of highly-dimensional computer vision applications, this approach allowed RL algorithms to obtain outstanding accuracy~\cite{grill2020bootstrap} in a variety of tasks and applications.

In medical imaging, model-free reinforcement learning algorithms are highly used for landmark detection~\cite{ghesu} as well as localization tasks~\cite{brest_lesion}. In~\cite{alansary2018automatic}, a Deep Q-Network (DQN) that automates the view planning process on brain and cardiac MRI was proposed. This framework takes as an input a single plane and updates its angle and position during the training process until convergence. Moreover, in~\cite{vlontzos2019multiple} the authors again present a DQN framework for the localization of different anatomical landmarks introducing multiple agents that act and learn simultaneously. DRL has also been studied for object or lesion localization. More recently, in~\cite{navarro2020deep} the authors propose a DQN framework for the localization of $6$ different organs from CT scans achieving a performance comparable to supervised CNNs. This framework uses a $3$D volume as input with $11$ different actions to generate bounding boxes for these organs. Our work is the first to explore and validate a RL scheme on MIP representations for a single slice detection using the discrete and 2D nature of the problem.

\section{Reinforcement Learning Strategy}
In this paper, we formulate the slice localization problem as a \textit{Markov Decision Process} (MDP), which contains a set of states $S$, actions $A$, and rewards $R$. 

\textbf{States $S$}: For our formulation, the environment $\mathcal{E}$ that we explore and exploit is a $2$D image representing the frontal MIP projection of the $3$D CT scans. This projection allows us to reduce our problem's dimensionality from a volume of size $512 \times 512 \times N$ ($N$ being the varying heights of the volumes) to an image of size $512 \times N$. The reinforcement learning agent is self-taught by interacting with this environment, executing a set of actions, and receiving a reward linked to the action taken. An input example is shown in Figure~\ref{fig:architecture}. We define a state $s \in S$ as an image of size $512 \times 200$ in $\mathcal{E}$. We consider the middle of the image to be the slice's current position on a $z$-axis. To highlight this, we assign a line of maximum intensity pixel value to the middle of each image provided as input to our DQN.

\textbf{Actions $A$}: We define a set of discrete actions $\mathcal{A} = \{ t^{+}_z, t^{-}_z \} \in \mathbb{R}^{2}$. $t^{+}_z$ corresponds to a positive translation (going up by one slice) and $t^{-}_z$ corresponds to a negative translation (going down by one slice). These two actions allow us to explore the entirety of our environment $\mathcal{E}$. 

\textbf{Rewards $R$}: In reinforcement learning designing a good reward function is crucial in learning the goal to achieve. To measure the quality of taking an action $a \in A$ we use the distance over $z$ between the current slice and the annotated slice $g$. The reward for non-terminating states is computed with:
\begin{equation}
 R_a(s,s')= sign(\mathcal{D}(p,g) - \mathcal{D}(p',g))
\end{equation}
where we denote as $s$ and $s'$ the current and next state and $g$ the ground truth annotation. Moreover, our positions $p$ and $p'$ are the $z$-coordinates of the current and next state respectively. $\mathcal{D}$ is the Euclidean distance between both coordinates over the $z$-axis. The reward is non-sparse and binary $r \in \{-1,+1\}$ and helps the agent differentiate between good and bad actions. A good action being when the agent gets closer to the correct slice. 
For a terminating state, we assign a reward of $r=0.5$.

\textbf{Starting States}: An episode starts by randomly sampling a slice over the $z$-axis and ends when the agent has achieved its goal of finding the right slice. The agent then executes a set of actions and collects rewards until the episode terminates. When reaching the upper or lower borders of an image, the current state is assigned to the next state (i.e., the agent does not move), and a reward of $r=-1$ is appointed to this action. 

\textbf{Final States}: During training, a terminal state is defined as a state in which the agent has reached the right slice. A reward of $r=0.5$ is assigned in this case, and an episode is terminated. 
During testing, the termination of an episode happens when oscillations occur. We adopted the same approach as~\cite{alansary2018automatic}, and chose actions with the lowest $Q$-value, which have been found to be closest to the right slice since the DQN outputs higher $Q$-values to actions when the current slice is far from the ground truth.

\subsection{Deep Q-Learning}
To find the optimal policy $\pi^{*}$ of the MDP, a state-action value function $Q(s,a)$ can be learned. In Q-Learning, the expected value of the accumulated discounted future rewards can be estimated recursively using the Bellman optimality equation:
\begin{equation}
 Q_{i+1}(s,a) = \mathbb{E}[ r + \gamma \underset{a^\prime}{\mathrm{max}}\:Q_{i}(s^\prime,a^\prime) \mid s,a]
\end{equation}

In practice, since the state $S$ is not easily exploitable, we can take advantage of neural networks as universal function approximators to approximate $Q(s,a)$~\cite{Mnih2015}. We utilize an experience replay technique that consists in storing the agent's experience $e_t= (s,a,r,s^\prime,a^\prime)$ at each time step in a replay memory $M$. To break the correlation between consecutive samples, we will uniformly batch a set of experiences from $M$. The Deep Q-Network (DQN) will iteratively optimize its parameters $\theta$ by minimizing the following loss function: 
\begin{equation}
 L_i(\theta_{i}) = \mathbb{E}_{(s,a,r,s^\prime,a^\prime)}\Big[ {\big( r + \gamma \underset{a'}{\mathrm{max}}\:Q_{target}(s^\prime,a^\prime;\theta_i^{-}) - Q_{policy}(s,a;\theta_{i}) \big)}^{2} \Big]
\end{equation}
with $\theta_{i}$ and $\theta_{i}^{-}$ being the parameters of the policy and the target network respectively.
To stabilize rapid policy changes due to the distribution of the data and the variations in Q-values, the DQN uses $Q_{target}(\theta_i^{-})$, a fixed version of $Q_{policy}(\theta_i)$ that is updated periodically. For our experiments, we update $\theta_{i}^{-}$ every 50 iterations.

\subsection{Network Architecture}
\begin{figure}[t!]
\includegraphics[width=\textwidth]{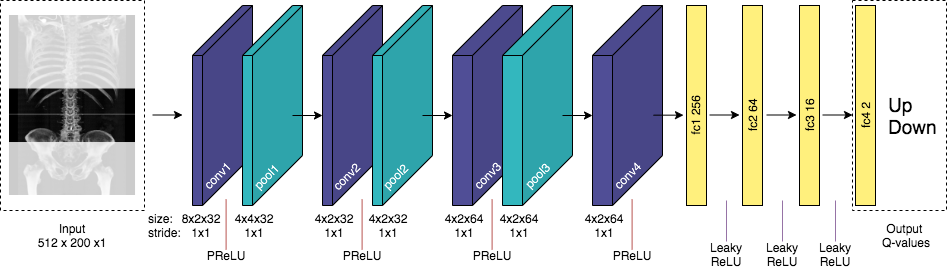}
\caption{The implemented Deep Q-Network architecture for L3 slice localization. The network takes as input an image of size $512 \times 200$ with a single channel. The output is the q-values corresponding to each of the two actions.} \label{fig:architecture}
\end{figure}

Our Deep Q-Network takes as input the state $s$ and passes it through a convolutional network. The network contains four convolution layers separated by parametric ReLU in order to break the linearity of the network, and four linear layers with LeakyReLU.
Contrary to~\cite{alansary2018automatic}, we chose not to add the history of previously visited states in our case. We opted for this approach since there is a single path that leads to the right slice. This approach allows us to simplify our problem even more. Ideally, our agent should learn, just by looking at the current state, whether to go up or down when the current slice is respectively below or above the L3 slice. An overview of our framework is presented in Figure~\ref{fig:architecture}.

We also explore dueling DQNs from~\cite{WangDueling}. Dueling DQNs rely on the concept of an advantage which calculates the benefit that each action can provide.
The advantage is defined as $A(s,a) = Q(s,a) - V(s)$ with $V(s)$ being our state value function. This algorithm will use the advantage of the Q-values to distinguish between actions from the state's baseline values. Dueling DQNs were shown to provide more robust agents that are wiser in choosing the next best action. For our dueling DQN, we use the same architecture as the one in Figure~\ref{fig:architecture} but change the second to last fully connected layer to compute state values on one side, and action values on the other.

\subsection{Training}
Since our agent is unaware of the possible states and rewards in  $\mathcal{E}$, the exploration step is implemented first. After a few iterations, our agent can start exploiting what it has learned on $\mathcal{E}$. 
In order to balance between exploration and exploitation, we use an $\epsilon$-greedy strategy. This strategy consists of defining an exploration rate $\epsilon$, which is initialed to $1$ with a decay of $0.1$, allowing the agent to become greedy and exploit the environment. 
A batch size of $48$ and an experience replay of $17 \times 10^3$ are used. 
The entire framework was developed in Pytorch~\cite{pytorch} library using an NVIDIA GTX 1080Ti GPU. We trained our model for $10^5$ episodes, requiring approximately $20$-$24$ hours. Our source code is available on GitHub: \url{https://git.io/JRyYw}.

\section{Experiments and Results}

\subsection{Dataset} 
A diverse dataset of $1000$ CT scans has been retrospectively collected for this study. CT scans were acquired on four different CT models from three manufacturers (Revolution HD from GE Healthcare, Milwaukee, WI; Brillance 16 from Philips Healthcare, Best, Netherlands; and Somatom AS+ \& Somatom Edge from Siemens Healthineer, Erlangen, Germany). Exams were either abdominal, thoracoabdominal, or thoraco-abdominopelvic CT scans acquired with or without contrast media injection. Images were reconstructed using abdominal kernel with either filtered back-projection or iterative reconstruction. Slice thickness ranged from $0.625$ to $3$mm, and the number of slices varied from $121$ to $1407$. The heterogeneity of our dataset highlights the challenges of the problem from a clinical perspective.

Experienced radiologists manually annotated the dataset, indicating the position of the middle of the L3 slice. Before computing the MIP, all of the CT scans are normalized to $1mm$ over the $z$-axis. This normalisation step harmonises our network's input, especially since the agent performs actions along the $z$-axis. After the MIP, we apply a threshold of $100$ HU (Hounsfield Unit) to $1500$ HU allowing us to eliminate artifacts and foreign metal bodies while keeping the skeleton structure. The MIP are finally normalized to [0,1]. From the entire dataset, we randomly selected $100$ patients for testing and the rest $900$ for training and validation. For the testing cohort, annotations of L3 from a second experienced radiologist have been provided to measure the interobserver performance.

\subsection{Results and Discussion}
Our method is compared with other techniques from the literature. The error is calculated as the distance in millimeters ($mm$) between the predicted L3 slice and the one annotated by the experts.  In particular, we performed experiments with the L3UNet-2D~\cite{kanavati2018automatic} approach and the winning  SC-Net~\cite{Payer2020} method of the Verse2020\footnote{\url{https://verse2020.grand-challenge.org/}} challenge. Even if SC-Net is trained on more accurate annotations with 3D landmarks as well as  vertebrae segmentations, and addresses a different problem, we applied it to our testing cohort. The comparison of the different methods is summarised in Table~\ref{tab1}. SC-Net reports $12$ CT scans with an error higher than $10$mm. Moreover, L3UNet-2D~\cite{kanavati2018automatic} reports a mean error of $4.24$mm $\pm6.97$mm when the method is trained on the entire training set, giving only $7$ scans with an error higher than $10$mm for the L3 detection. Our proposed method gives the lowest errors with a mean error of $3.77$mm$\pm4.71$mm, proving its superiority.
Finally, we evaluated our technique's performance with a Duel DQN strategy, reporting higher errors than the proposed one. This observation could be linked to the small action space that is designed for this study. Duel DQNs were proven to be powerful in cases with higher action spaces and in which the computation of the advantage function makes a difference.

\begin{table}[t!]
\caption{Quantitative evaluation of the different methods using different number of training samples (metrics in mm).}\label{tab1}
\begin{center}
\begin{tabular}{|l|l|l|l|l|l|l|}
\hline
Method & \# of samples & Mean & Std & Median & Max & Error $> 10mm$\\
\hline
Interobserver & - & 2.04 & 4.36 &  1.30 & 43.19 &  1\\ \hline  \hline
SC-Net~\cite{Payer2020} & - & 6.78 & 13.96 &  {\bfseries 1.77} & 46.98 &  12\\ \hline 
L3UNet-2D~\cite{kanavati2018automatic} & 900 & 4.24 & 6.97 & 2.19 & 40 & {\bfseries 7}\\
Ours (Duel-DQN) & 900 & 4.30 &  5.59 &  3 &  38 & 8\\
Ours & 900 & {\bfseries 3.77} & {\bfseries 4.71} &  2.0 & {\bfseries 24} & 9\\
\hline  \hline
L3UNet-2D~\cite{kanavati2018automatic}  & 100 & 145.37 & 161.91 & 32.8 & 493 & 68\\
Ours & 100 & {\bfseries 5.65} & {\bfseries 5.83} & {\bfseries 4} & {\bfseries 26} & {\bfseries 19}\\ \hline  \hline
L3UNet-2D~\cite{kanavati2018automatic}  & 50 & 108.7 & 97.33 & 87.35 & 392.02 & 86\\
Ours & 50 & {\bfseries 6.88} & {\bfseries 5.79} & {\bfseries 6.5} & {\bfseries 26} & {\bfseries 11}\\ \hline \hline
L3UNet-2D~\cite{kanavati2018automatic} & 10 & 242.85 & 73.07 & 240.5 & 462 & 99\\
Ours & 10 & {\bfseries 8.97} & {\bfseries 8.72} & {\bfseries 7} & {\bfseries 56} & {\bfseries 33}\\
\hline
\end{tabular}
\end{center}
\end{table}

For the proposed reinforcement learning framework, trained on the whole training set, 9 CTs had a detection error of more than $10$mm. These scans were analysed by a medical expert who indicated that $2$ of them have a lumbosacral transitional vertebrae (LSTV) anomaly~\cite{definitionlstv}. Transitional vertebrae cases are common and observed in 15-35\% of the population~\cite{lstv} highlighting once again the challenges of this task. 
For both cases, the localization of the L3 vertebra for sarcopenia assessment is ambiguous for radiologists and consequently for the reinforcement learning agent. In fact, the only error higher than $10mm$ in the interobserver comparison corresponds to an LSTV case where each radiologist chose a different vertebrae as a basis for sarcopenia assessment. Even if the interobserver performance is better than the one reported by the algorithms, our method reports the lowest errors, proving its potential.

\begin{figure}[t!]
   \includegraphics[width=1.0\textwidth]{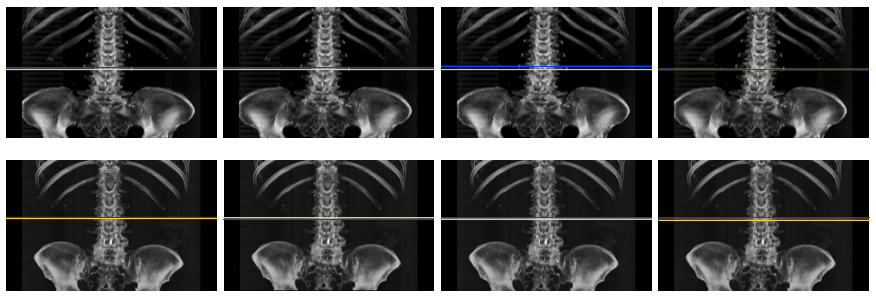}
   \centering
   \caption{Qualitative comparison of different localization methods for two patients. First left to right represents: interobserver ($4mm$/ $2mm$), SC-Net ($4mm$/$2mm$), L3UNet-2D ($8mm$/$1mm$), Ours ($1mm$/$4mm$). In the parenthesis we present the reported errors for the first and second row respectively.
   The yellow line represents the ground truth and the blue one the prediction.} 
\label{fig:comparison}
\end{figure}

Qualitative results are displayed in Figure~\ref{fig:comparison}. The yellow line represents the medical expert's annotation and the blue one the prediction of the different employed models. One can notice that all of the different methods converge to the correct L3 region with our method reporting great performance. It is important to note that for sarcopenia assessment, an automatic system does not need to be at the exact middle of the slice; a few millimeters around will not skew the end result since muscle mass in the L3 zone does not change significantly. Concerning prediction times for the RL agent, they depend on the initial slice that is randomly sampled on the MIP. Computed inference time for a single step is approximately 0.03 seconds.

To highlight the robustness of our network on a low number of annotated samples, we performed different experiments using 100, 50, and 10 CTs corresponding respectively to 10\%, 5\% and 1\% of our dataset. We tested those 3 agents on the same 100 patients test set and report results in Table~\ref{tab1}. Our experiments prove the robustness of reinforcement learning algorithms compared to traditional CNN based ones~\cite{kanavati2018automatic} in the case of small annotated datasets. One can observe that the traditional methods fail to be trained properly with a small number of annotations, reporting errors higher than $100$mm for all three experiments. In our case, decreasing the dataset size does not significantly affect the performance. In fact, trained on only 10 CTs with the same number of iterations and memory size, our agent was able to learn a correct policy and achieve a mean error of $8.97$mm $\pm$ $8.72$mm. Learning a valid policy from a low number of annotations is one of the strengths of reinforcement learning. Traditional deep learning techniques rely on pairs of images and annotations in order to build a robust generalization. Thus, each pair is exploited only once by the learning algorithm. Reinforcement learning, however, relies on experiences, each experience $e_t= (s,a,r,s^\prime,a^\prime)$ being a tuple of state, action, reward, next state and next action. Therefore, a single CT scan can provide multiple experiences to the self-learning agent, making our method ideal for slice localization problems using datasets with a limited amount of annotations.

\section{Conclusion}
In this paper, we propose a novel direction to address the problem of CT slice localization. Our experiments empirically prove that reinforcement learning schemes work very well on small datasets and boost performance compared to classical convolutional architectures. One limitation of our work lies in the fact that our agent is always moving $1mm$ independently of the location, slowing down the process. In the future, we aim to explore different ways to adapt the action taken depending on the current location, with one possibility being to incentivize actions with higher increments. Future work also includes the use of reinforcement learning in multiple vertebrae detection with competitive or collaborative agents. 
%
%
%
 \bibliographystyle{splncs04}
 \bibliography{bib}
\end{document}